\documentclass[9pt,conference]{IEEEtran}
\IEEEoverridecommandlockouts
\usepackage{subcaption}
\usepackage{cite}
\usepackage{amsmath,amssymb,amsfonts}
\usepackage{algorithmic}
\usepackage{graphicx}
\usepackage{textcomp}
\usepackage{xcolor}
\usepackage{soul}
\def\BibTeX{{\rm B\kern-.05em{\sc i\kern-.025em b}\kern-.08em
    T\kern-.1667em\lower.7ex\hbox{E}\kern-.125emX}}

\makeatletter
\def\footnoterule{\relax%
  \kern-5pt
  \hbox to \columnwidth{\hfill\vrule width 1\columnwidth height 0.4pt\hfill}
  \kern4.6pt}
\makeatother

\definecolor{ilias_color_TM}{RGB}{191, 232, 255}
\definecolor{stelios_colour}{RGB}{144, 238, 144}
\definecolor{giannis_colour}{RGB}{191, 132, 155}

\newif\ifcomment

\commenttrue

\ifcomment
\newcommand{\stelios}[1]{\sethlcolor{stelios_colour}\hl{[\textbf{Stelios:} #1]}}
\newcommand{\il}[1]{\sethlcolor{ilias_color_TM}\hl{[\textbf{Ilias:} #1]}}
\newcommand{\giannis}[1]{\sethlcolor{orange}\hl{[\textbf{Giannis:} #1]}}
\newcommand{\cut}[1]{\sethlcolor{light_red}\hl{[#1]}}
\else
\newcommand{\stelios}[1]{}
\newcommand{\il}[1]{}
\newcommand{\giannis}[1]{}
\newcommand{\cut}[1]{}

\fi

\usepackage{background}
\backgroundsetup{angle=0,
    scale=1,
    color=black,
    firstpage=true,
    position=current page.north,
    hshift=0pt,
    vshift=-20pt,
    contents={\ifnum\value{page}=1 PREPRINT: Invited paper at the 32nd IEEE International Conference on Application-Specific Systems, Architectures and Processors (ASAP), 2021 \else \fi}
}

\begin{document}

\title{How to Reach Real-Time AI on \mbox{Consumer Devices?}\\  \mbox{Solutions for Programmable and Custom Architectures}}
\IEEEspecialpapernotice{(Invited Paper)}

\author{\IEEEauthorblockN{
Stylianos I. Venieris\IEEEauthorrefmark{2},   
Ioannis Panopoulos\IEEEauthorrefmark{3},
Ilias Leontiadis\IEEEauthorrefmark{2},
Iakovos S. Venieris\IEEEauthorrefmark{3}
}
\IEEEauthorblockA{\IEEEauthorrefmark{2}Samsung AI Center, Cambridge, UK,
\IEEEauthorrefmark{3}National Technical University of Athens, Athens, Greece}
\vspace{-0.8cm}
}

\maketitle

\begin{abstract}
The unprecedented performance of deep neural networks (DNNs) has led to large strides in various Artificial Intelligence (AI) inference tasks, such as object and speech recognition. Nevertheless, deploying such AI models across commodity devices faces significant challenges: large computational cost, multiple performance objectives, hardware heterogeneity and a common need for high accuracy, together pose critical problems to the deployment of DNNs across the various embedded and mobile devices in the wild. As such, we have yet to witness the mainstream usage of state-of-the-art deep learning algorithms across consumer devices.
In this paper, we provide preliminary answers to this potentially game-changing question by presenting an array of design techniques for efficient AI systems. We start by examining the major roadblocks when targeting both programmable processors and custom accelerators. Then, we present diverse methods for achieving real-time performance following a cross-stack approach. These span model-, system- and hardware-level techniques, and their combination. Our findings provide illustrative examples of AI systems that do not overburden mobile hardware, while also indicating how they can improve inference accuracy. Moreover, we showcase how custom ASIC- and FPGA-based accelerators can be an enabling factor for next-generation AI applications, such as multi-DNN systems. Collectively, these results highlight the critical need for further exploration as to how the various cross-stack solutions can be best combined in order to bring the latest advances in deep learning close to users, in a robust and efficient manner.
\end{abstract}

\vspace{-0.1cm}
\section{Introduction}
\label{sec:intro}

The unprecedented predictive power of deep neural networks (DNNs) has led to their ever-increasing usage on mobile and embedded devices, transforming their capabilities and, consequently, our lives. At the same time, \textit{real-time} AI applications are also gaining popularity. 
For instance, smart assistants are required to respond with low latency~\cite{ondevice_asr2020getmobile} while AI video upscaling algorithms are required to run at high frame rates in order to avoid rebuffering~\cite{nvidia_dynamic_sr,neural_enhancement2021csur}.

Supporting  real-time requirements on mobile hardware is becoming more and more challenging as the complexity of state-of-the-art DNNs is increasing exponentially~\cite{complexity}.
Most device vendors have started incorporating System-on-Chips (SoCs) that can accelerate DNN computations, such as GPUs and NPUs. While these can significantly accelerate DNN inference,   developers still face the same issue: \emph{they have to support the wide variety of devices} that can be found in the wild~\cite{embench2019emdl,ai_benchmark2019iccvw,fb_edge2019hpca}. This includes older devices, low- and mid-range smartphones, wearables and IoT devices. Hence, developers frequently resort to deploying simpler or heavily compressed CNNs at the expense of accuracy~\cite{pruning2020mlsys}. 
As real-time inference is not always possible, DNN developers also rely on fully or partially offloading to a remote infrastructure, 
such as the cloud or the edge~\cite{neurosurgeon2017asplos,spinn2020mobicom}. 
Offloading can improve inference latency and resolve the problem of wide device compatibility, at the expense of using network and cloud resources, raising privacy concerns~\cite{darknetz2020mobisys} 
and yielding inconsistent user experience due to varying networking conditions~\cite{adaptive_mco_2012cloudnet}.

While on-device optimisations and computation offloading can help supporting some real-time requirements,  upcoming  applications impose even stricter deadlines: self-driving cars need to process multi-sensor inputs within a few hundred milliseconds~\cite{self_driving_cars2018asplos}, an AR/VR headset typically performs scene recognition within 20 milliseconds while supporting 120 Hz frame-rates~\cite{vr_latency2018mobisys}, whereas robotic assistants need to run multiple models simultaneously to achieve context awareness and to interact with their environment. Typically, these scenarios are only addressed by co-designing DNNs with domain-specific hardware, such as ASICs and FPGA-based accelerators.

In this paper, we will dive into prominent techniques that have been used to support real-time AI  in both \emph{general-purpose} and \emph{customised} hardware platforms.  We start by examining  the  major  roadblocks and then present  diverse methods for achieving real-time performance that span the whole stack: model-, system- and  hardware-level techniques, and their combination. Moreover, we showcase how and under which settings custom ASIC- and FPGA-based accelerators can be an enabling factor for next-generation AI applications.


\vspace{-0.1cm}
\section{Common Roadblocks in Real-Time AI}
\label{sec:roadblocks}

In an AI system, a stream of input samples (\textit{e.g.}~photos, video frames, mic signals, or accelerometer readings) is processed by an AI model, typically a DNN, in order to perform an inference (\textit{e.g.}~object or speech recognition). Central to the operation of such a system is the hardware processing unit that executes the DNN inference. The architectural landscape of processing units for AI workloads can be classified into two main categories: \textit{1)}~programmable processors and \textit{2)}~custom accelerators. This classification is based on the \textit{efficiency-flexibility trade-off} of the underlying hardware.

Despite the radical progress of deep learning, only a few big vendors have been in position to integrate state-of-the-art AI technologies across all their products. Even in these cases, a number of critical issues are challenging the efficient and wide integration of DNN-based algorithms in consumer devices:

\subsubsection{\textbf{DNN Diversity}} DNN models vary in terms of task, architecture, workload and resource demands. These factors have a direct impact on the memory footprint, number of operations, computation-to-communication ratio, the parallelisation potential and the resilience to approximate computing techniques~\cite{approx_hw2019csur}.

For classification tasks, even from 2012, DNNs such as AlexNet and VGG-16 exhibited orders of magnitude higher computational demands than other ML models. This was further aggravated with the development of large-scale models, such as ResNet-152 and DenseNet-161. Despite the design of efficient models, such as MobileNet and ShuffleNet, that employ novel blocks, such as depthwise separable convolutions, to reduce the number of operations, these blocks are often memory-bounded or underutilise the underlying processing hardware. As such, the theoretical complexity reduction does not always translate to actual performance gains upon deployment.

At the same time, tasks such as image/video super-resolution~\cite{mobisr2019mobicom} and semantic segmentation~\cite{mess2021arxiv}, are characterised by even larger computational complexity. This mainly stems from the fact that, in contrast to classification DNNs that reduce the feature maps' size as we go deeper in the network, these tasks require the size of the feature maps to be maintained. The rationale behind this is that high-quality super-resolution or segmentation require the propagation of information about high-frequency details, such as the texture or the contour of an object, until the output of the DNN. This property affects significantly both the memory footprint and the number of operations, imposing a barrier in achieving real-time performance.

In the field of NLP and ASR, applications are dominated by RNNs (\textit{e.g.}~LSTMs/GRUs) and Transformers. The primary computational challenge of these families of DNNs is that they consist of multiple matrix-vector multiplications and hence are memory-bounded.
As a result, processors and accelerators that have typically been optimised for compute-bound convolutional layers and matrix-matrix multiplications are pushed to their limits~\cite{tpu2017isca} and performance becomes bounded by the available off-chip memory bandwidth~\cite{brainwave2018isca}. The same holds for the case of Multi-Layer Perceptrons (MLPs) that rely only on the memory-bound fully-connected (FC) layers~\cite{tpu2017isca}.

Recently, neural architecture search (NAS) methodologies~\cite{nas2017iclr} have rapidly been adopted to automatically generate highly accurate and, sometimes, compact models for a target task. Nonetheless, NAS often leads to nonintuitive topologies, up to the extreme case of randomly wired networks~\cite{rwn2019iccv,rwn2019neurips}. The complex and irregular topology of such DNNs poses important problems in terms of both compiling them for existing programmable processors~\cite{rwn2020mlsys} and deriving a suitable custom accelerator~\cite{rwn2020fpl}.

In this context, the rapid algorithmic advancements from the AI community are in need for future-proof solutions and hence call for \textit{general} hardware platforms that can be re-used from the following generations of DNNs. On the other hand, high performance often requires customisation, which in turn hurts generality. As a result, finding a balance between flexibility and customisation remains a challenging and crucial problem in the design of AI hardware.



\subsubsection{\textbf{Performance Objectives' Variability}}
Depending on the end application and target device, the performance requirements vary significantly in terms of accuracy, latency, throughput, energy and power across DNN applications. Even under the unified goal of real-time processing, the application determines the lowest acceptable accuracy and the platform dictates the available energy, power and resource budget of the system. For instance, interactive applications, such as VR and gaming, demand low latency (\textit{e.g.}~20 ms), while wearable devices require ultra-low-power solutions (\textit{e.g.}~$<$1 W).

\subsubsection{\textbf{System Heterogeneity}}
The different processing capabilities of devices in the wild lead to wide system heterogeneity. This comprises both the system software and the underlying hardware. On the software side, the fragmented space of OS variants (\textit{e.g.}~numerous versions of Android, iOS, Tizen, \textit{etc}), together with the partial support of a unified middleware (\textit{e.g.}~limited support and inconsistent performance of {\texttt{NNAPI}} across smartphones~\cite{fb_edge2019hpca,ai_benchmark2019iccvw,oodin2021smartcomp}), poses challenges in maintaining the functionality and performance through time and across devices. On the hardware side, the large number of vendors and the different use-cases have led to devices with broadly different characteristics~\cite{fb_edge2019hpca,ai_benchmark2019iccvw,embench2019emdl,oodin2021smartcomp}, such as processing capabilities, memory capacity, camera, mic and accelerometer sensors.
As a result, performance cannot be trivially sustained across devices, leading to inconsistent quality of experience (QoE) for users of different devices. 

\subsubsection{\textbf{Environment Dynamicity}}
Dynamicity is often manifested in the form of reduced processing speed, longer delays during memory transfers and degraded network bandwidth. The roots of this phenomenon stem from \textit{i)}~the multi-tasking nature of mobile systems~\cite{nestdnn2018mobicom}, \textit{ii)}~the frequency throttling policies that are in-place to avoid overheating~\cite{dvfs2020mdat} and \textit{iii)}~the fluctuations in the quality of the network connectivity~\cite{realitycheck2019edgesys}. These factors often make the static design analysis and performance estimation futile, and necessitate the design of systems that can dynamically adapt to changes.

\section{Real-Time AI on Programmable Processors}
\label{sec:rt-ai-prog-archs}

Consumer devices, such as smartphones and tablets, typically host processors that are able to serve a multitude of diverse workloads. As such, their design follows a more general-purpose approach and favours flexibility and programmability. We define as programmable processor any architecture that consists of processing elements that execute a stream of instructions, \textit{without introducing domain-specific optimisations at the hardware or ISA level}.

Such processors span from ubiquitous mobile CPUs, such as Arm Cortex-A, Qualcom Kryo and Samsung Exynos~\cite{exynos_cpu2020isca}, up to more specialised units, such as mobile GPUs, DSPs and NPUs. This class of processors can be found in many flavours, based on the performance needs of the application and the cost, power and form-factor constraints of the target platform. 
For instance, flagship smartphones tend to host more powerful CPUs (\textit{e.g.}~the Arm Cortex-X1 core in Samsung S21 Ultra) and GPUs than their mid- (\textit{e.g.}~Kryo 400 series in Samsung Galaxy A72) and low-tier (\textit{e.g.}~Arm Cortex-53 in Samsung Galaxy J7) counterparts. A similar situation can be observed for notebook and tablets which can host powerful processors with a medium power limit (\textit{e.g.}~Apple M1 on MacBook and iPad Pro with 15-watt TDP) compared to phones with tighter thermal limits (\textit{e.g.}~Apple A14 SoC with 5 TDP on iPhone 12).
On the other hand, IoT devices, such as smart watches and home sensors, often rely on energy-efficient, but memory-constrained, microcontrollers (MCUs), so that they can be unintrusively integrated into the users' everyday life. 
Nonetheless, the extensive flexibility of programmable architectures comes at the cost of a hard limit on the attainable processing speed and energy efficiency~\cite{proc_inefficiencies2010isca}.

In the rest of this section, we present an array of solutions that make important strides towards real-time AI, highlighting the essential components to achieve this goal.
We classify these solutions based on the entity of the system where the optimisation is implemented:
\begin{itemize}
    \item System optimisations (Section~\ref{sec:prog-arch-sys})
    \item Model optimisations (Section~\ref{sec:prog-arch-model})
    \item Joint model-system optimisations (Section~\ref{sec:prog-arch-model-sys})
\end{itemize}

\subsection{System Optimisations}
\label{sec:prog-arch-sys}

\begin{figure}[t]
    \begin{subfigure}{.24\textwidth}
        \centering
        \includegraphics[width=1\textwidth,trim={0cm 11cm 23.5cm 0cm},clip]{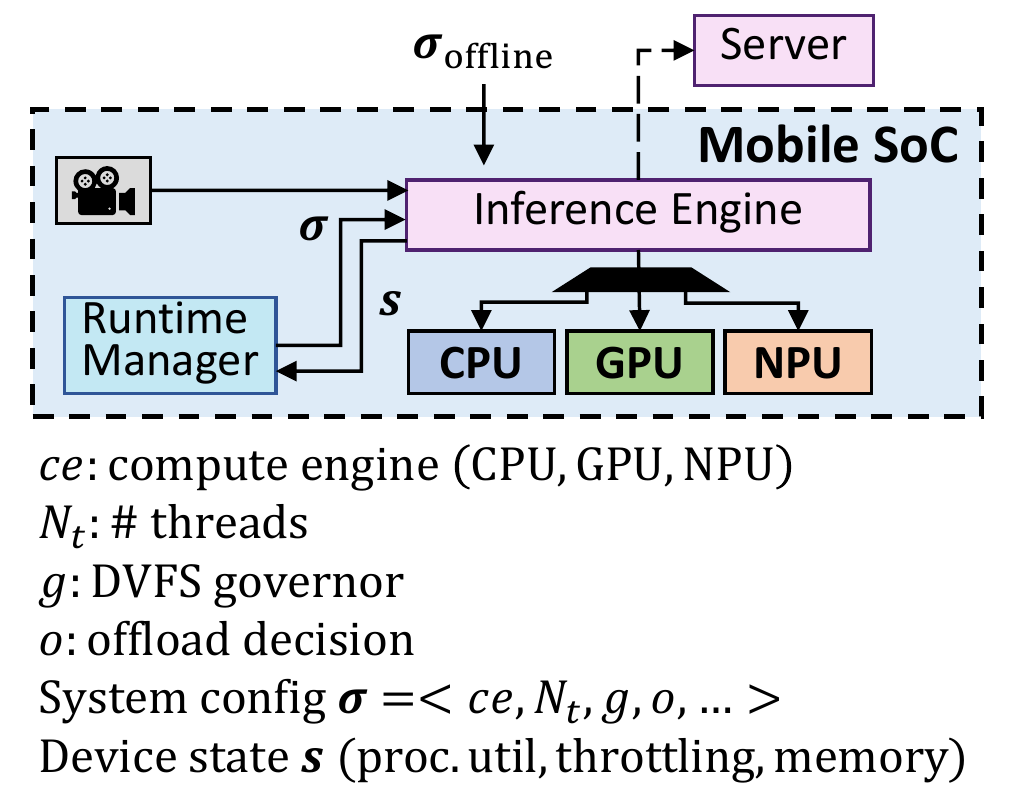}
        \caption{}
        \label{fig:ondevice-sys-opt}
    \end{subfigure}
    \begin{subfigure}{.24\textwidth}
        \centering
        \vspace{1.8cm}
        \includegraphics[width=1\textwidth,trim={0cm 7cm 18.5cm 6.5cm},clip]{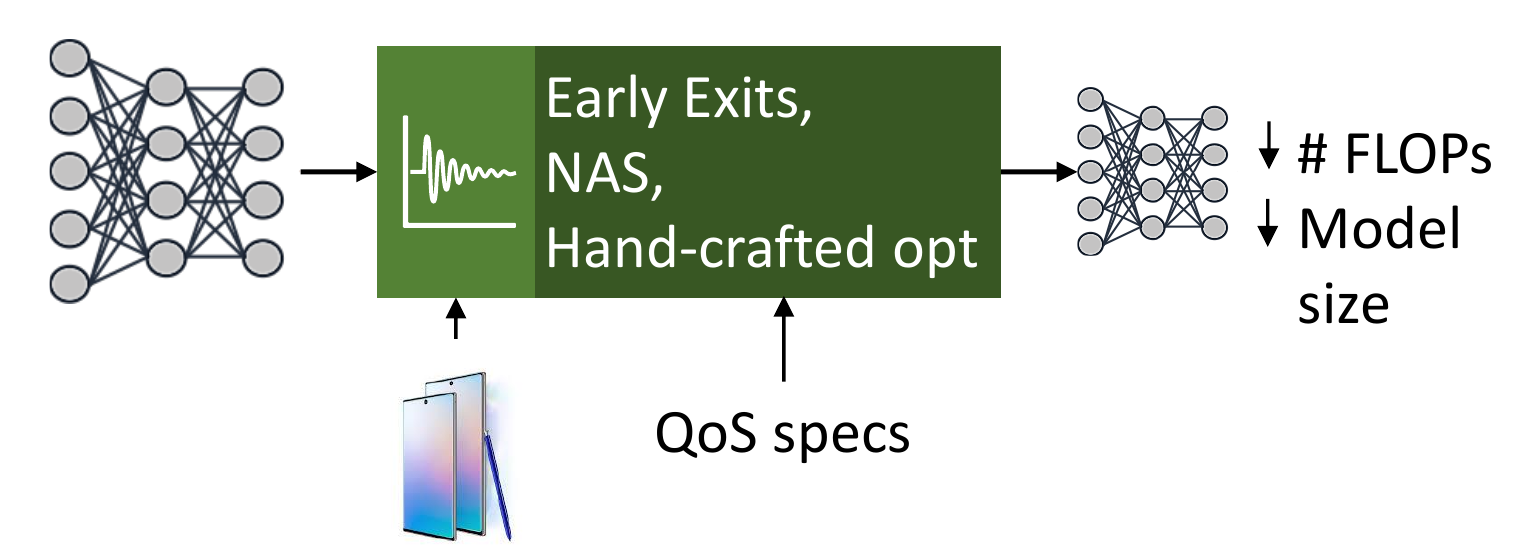}
        \caption{}
        \label{fig:hw-model-adapt}
    \end{subfigure}
    \caption{System tuning (a) and model adaptation~(b), performed in isolation or jointly, constitute pillars of real-time AI.
    }
    \vspace{-0.5cm}
\end{figure}

An approach of addressing the system heterogeneity and meeting real-time performance for AI inference is to adapt the deployment to the characteristics of the device at hand. This adaptation process involves finding the highest-performing resource configuration of the target mobile SoC, such as enabling and disabling cores of different types, defining the task-to-processor mapping, setting the dynamic voltage and frequency scaling (DVFS) policy and making server offloading decisions (Fig.~\ref{fig:ondevice-sys-opt}).
With the exception of dynamic DNNs~\cite{nimble2021mlsys,adaptive_dnns2021emdl}, the majority of deep learning models are characterised by a static workload which is known before run time. This advocates for an initial \textit{static optimisation stage}. At the same time, modern consumer devices are increasingly dealing with concurrent execution of apps with various resource demands, performance needs and random arrival/completion times. As such, \textit{dynamic adaptation} mechanisms are also key behind sustaining the required performance during DNN inference.


\textbf{Static \& Dynamic System Adaptation:}
OODIn~\cite{oodin2021smartcomp} is an on-device framework that showcases the potential of system tuning to tailor the DNN inference to the target platform. To capture the multiple objectives of DNN inference workloads, OODIn introduces a multi-objective optimisation framework that combines resource constraints with accuracy and performance requirements.
Next, the framework identifies key system parameters, including the task-to-processor mapping, the number of threads, the DVFS policy and the level of precision quantisation of the DNN model, and exposes them for optimisation to tailor the execution of the DNN to both the application-level performance needs and the underlying hardware. 
To this end, OODIn’s workflow is divided into two components: the offline (or static) and the online (or dynamic).


During the offline stage, OODIn creates multiple model variants with different levels of quantisation in order to modify the accuracy-complexity trade-off of the user-supplied model. 
As such, OODIn's offline optimisation method takes into account both the model space and the user-supplied performance goals to yield the optimal model and system configuration. Static optimisation leads to average speedups of more than 70\% over highly optimised status-quo implementations across diverse devices and DNN models.

On the other hand, the online phase is responsible for the mobile application’s robustness and adaptability. OODIn tracks the mobile device’s dynamic recourse availability changes, due to multi-tasking or thermal throttling, and reconfigures the selected parameters. Timely and efficient dynamic adaptation leads to latency reductions of up to 2.7$\times$ over statically optimised configurations.

\textbf{Dynamic Onloading/Offloading:}
DNN developers who seek state-of-the-art performance and broad device compatibility, typically resort into offloading computation to a remote server, either on the cloud or the edge. While this can resolve the problem of supporting devices of various capabilities, cloud offloading can also result in high operation costs, privacy issues and excessive dependence on the networking conditions.

\textit{Computation onloading}~\cite{dyno2021arxiv} aims to combine the best of both worlds: i)~the cloud’s elastic computational power and the ability to support a wide variety of devices and ii)~the fact that modern embedded devices have, ever-increasing, DNN processing capabilities. The main idea is to split a DNN into two parts; during inference the device executes a part of the computation then transfers a heavily compressed version of the intermediate results to a powerful server to resume computation and then retrieve back the result. The main idea is to \emph{onload as much computation as possible from cloud-native models into resource constrained devices} in order to maximise the overall performance and reduce the cloud cost, while meeting the application deadlines. As a result, powerful devices can process most of the DNNs locally and, therefore, save cloud resources, whereas less powerful devices might need more support from a server. These systems monitor and dynamically adjust the split point at run time, automatically freeing resources from the cloud by dynamically utilising on-device hardware. Results show that dynamic onloading can lead to an order of magnitude higher inference throughput while saving cloud resources.

\subsection{Model Optimisations}
\label{sec:prog-arch-model}

Under settings where the underlying processing engine is assumed to be fixed, applying optimisations at the model level can lead to substantial gains (Fig.~\ref{fig:hw-model-adapt}). Well-investigated methods of reducing the cost of inference include quantisation~\cite{integer_dnn2018cvpr}, pruning~\cite{netadapt2018eccv} and low-rank factorisation~\cite{approx_lstms2020cemag}. Orthogonal to these methods, two prominent types of model optimisations that further push the performance on commodity processors constitute: \textit{1)}~hardware-aware model adaptation and \textit{2)}~hardware-agnostic efficient model design. Primary assumption in both cases is the availability of the training dataset for the target AI task, which enables the model-level modifications.

\subsubsection{\textbf{Hardware-aware Model Adaptation}}
\label{sec:adaptation}
Recently, a plethora of adaptive DNN architectures have been proposed. The overarching objective is to exploit the \textit{variability in complexity} of different input samples in order to perform only the necessary amount of computation to obtain an accurate prediction. Moreover, this class of DNNs can tunably scale their resource usage and thus \textit{dynamically adapt} to any fluctuations in resource availability, either due to thermal throttling or multi-tasking.
To this end, various input-dependent execution mechanisms have been proposed, leading to dynamic, conditional DNN models. Such mechanisms include dynamically pruned DNNs~\cite{dyn_channel_pruning2018iclr} and early-exit models~\cite{hapi2020iccad}.

\textbf{Hardware-aware Early-Exit DNNs:}
To extract peak performance, a stream of works has presented hardware-aware methods for the construction of early-exit DNNs~\cite{adaptive_dnns2021emdl}. Such frameworks consider the computational, memory and energy budget of a target platform, in order to strategically attach early exits across the depth of a given model and tune the associated early-exit policy.

HAPI~\cite{hapi2020iccad} is a representative model-adaptation framework whose goal is to convert vanilla DNNs into high-performance early-exit models. This is achieved through a hardware-aware methodology that considers both the characteristics of the target platform and the maximum latency tolerance in order to automatically select the number and position of early exits along the DNN architecture. As such, the early-exit DNN topology is \textit{statically} optimised before deployment. Then, \textit{at run time}, HAPI adopts a tunable confidence-based early-exiting policy which dictates that a sample will stop at the first exit that yields a confident-enough prediction. Through this fine-grained parametrisation, HAPI tailors the early-exit model (number and placement of exits) and the early-exit policy (confidence threshold) to both the app-level performance requirements and the platform capabilities, resulting in 2.33$\times$ speedup and 2.53 percentage points (pp) higher accuracy than MobileNetV2 on Nvidia Jetson Xavier under the same \mbox{10-watt} power budget, highlighting the gains that can be obtained through hardware-aware model adaptation.

\subsubsection{\textbf{Efficient Model Design}}
A promising approach that emphasises generality is the manual or automated design of efficient, lightweight models. Flows for efficient model design typically rely on platform-agnostic metrics, such as FLOP count and model size, to set a computational and memory budget. Although such proxy metrics often do not translate to actual processing gains~\cite{embench2019emdl,brpnas2020neurips}, notable performance gains have been achieved and mobile-friendly DNNs such as MobileNet~\cite{mobilenetv2_2018cvpr}, SqueezeNet~\cite{squeezenet2016arxiv} and EfficientNet~\cite{efficientnet2019icml} have been widely adopted in actual applications. Here, we describe three prominent approaches for designing efficient models.

\textbf{Budget-aware Neural Architecture Search:}
Recently, significant effort has been placed into NAS (or AutoML) frameworks that aim to find high-accuracy models under computational or memory constraints~\cite{morphnet2018cvpr,brpnas2020neurips}. These frameworks typically adopt device-independent metrics to guide their search towards compact models that would potentially meet the required performance \textit{across devices}.

Such a NAS-generated model is TPSR~\cite{tpsr2020eccv}, a compact DNN for the task of image super-resolution. Optimised for perceptual quality and small footprint, TPSR delivers high-quality $\times$4 image upscaling while consuming only 244 KB (FP32) or 61 KB (INT8) of memory. With an average latency of 71 ms per image (\textit{i.e.}~14 frames-per-second) when upscaling to 720p using the NPU of a Qualcomm Snapdragon 865 SoC, TPSR showcases the potential of budgeted NAS even for the challenging case of mapping expensive tasks on smartphones and other resource-constrained IoT platforms.

\textbf{AutoML-powered Model Compression:}
A drawback of running a complete NAS is the excessive computational requirement during the search phase.
To alleviate this cost, it is possible to parametrise existing DNNs with parameters that expose an accuracy-complexity trade-off and exploit the efficacy of AutoML in order to find a high-performing configuration for these values. An example of this is ShrinkML~\cite{shrinkml2019interspeech,compressed_asr2020interspeech} which targets streaming LSTM-based models for automatic speech recognition (ASR) on mobile devices. ShrinkML employs low-rank factorisation of each layer in order to tunably prune the DNN weights. Each layer is compressed down to a different degree, with the per-layer compression ratio determined automatically using a reinforcement learning-based AutoML controller. This leads to a 17 ms latency on an Exynos 9810 CPU, corresponding to 3$\times$ speedup over the vanilla model.

\textbf{Hand-crafted Model Optimisation:}
A third approach for achieving real-time performance is to apply hand-engineered optimisations. Typically, such techniques are designed by domain experts and exploit domain-specific opportunities to improve the attainable performance. An instance of such a technique can be observed in the design of the bunched-LPCNet model~\cite{bunched_tts2020intespeech} for Text-to-Speech (TTS) applications. The vanilla LPCNet is enhanced with \textit{sample bunching}, a technique that allows it to produce more than one sample per inference and, in turn, reduce the overall computational cost. This is achieved by grouping together $S$ temporally neighbouring samples and modifying the DNN architecture so that it can process all $S$ samples as a bunch.
Deployed on an Exynos 9820 CPU, bunched-LPCNet delivers a speedup of 2.19$\times$ over the non-optimised model and achieves a real-time factor of 0.137. As such, by exploiting both the temporal nature of TTS and the large capacity of the LPCNet's GRUs, bunched-LPCNet demonstrates the gains that can be obtained through careful hand-crafted optimisations.

\subsection{Joint Model-System Optimisation}
\label{sec:prog-arch-model-sys}

A key approach to further boost the attainable performance is the joint optimisation of both the model architecture and the system parameters. Such schemes encompass techniques such as using alternative convolutional layers that map efficiently on the target hardware, designing multiple models and intelligently scheduling each input sample on the most suitable one based on a criterion, and strategically parallelising across the various processors of modern mobile SoCs.

\textbf{Model Selection \& Heterogeneous Computing:}
MobiSR~\cite{mobisr2019mobicom}, a framework for efficient super-resolution on smartphones, exemplifies the merits of model-system co-design. With super-resolution DNNs being especially computationally demanding, the proposed system introduces optimisations at various levels: exploiting the difference in upscaling difficulty among the different patches of an image, MobiSR uses a pair of models, each pinned to a different processor of the phone. The architecture of each model is optimised to yield efficient execution on the associated processor. At run time, each image patch's difficulty is quantified based on a total-variation metric and scheduled to the appropriate model-processor pair. Through this model-system co-optimisation, MobiSR delivers 4.79$\times$ speedup over highly optimised single-processor implementations on a phone equipped with a Qualcomm Snapdragon 845 SoC.

\textbf{Offloading Early-Exit DNNs for Robust Inference:}
Another approach that aims at both high performance and robust inference when the connectivity of the device to a server is uncertain is presented by SPINN~\cite{spinn2020mobicom}. SPINN combines distributed device-server inference with early-exit DNNs to deliver fast and robust inference across dynamic settings. The proposed system jointly and dynamically optimises the early-exit policy of the DNN (model-level optimisation) and the device-server partition point (system-level optimisation), providing previously unattainable adaptability to dynamic conditions. As such, SPINN achieves 2$\times$ higher throughput over existing distributed inference systems that solely optimise system parameters. Moreover, by always placing an early exit on the device, the accuracy is maintained high even under severely constrained server availability. The concurrent use of distributed execution, adaptive early-exit DNNs and run-time system tuning leads to new levels of flexibility and enables deployment across diverse devices.

\subsection{How personalised DNNs can help?}
\label{sec:prog-arch-personal}

To be deployable in the wild, AI models need to generalise across a wide variety of inputs.
For instance, facial landmark detectors are trained to capture various demographics, speech recognisers to accommodate different accents and voices, and home assistant robots to work reliably across diverse household configurations.
Traditionally, to handle all these scenarios, parameter-heavy and computationally costly models are trained on massive datasets that aim to capture the majority of cases that will be encountered upon deployment. 
In contrast to this approach, a different paradigm introduces \textit{on-device model personalisation}, aiming to tailor the DNN to a specific user or environment. Personalised models can be used not only to improve accuracy, but also as a way to improve efficiency.

One way to improve efficiency is to personalise early-exit DNNs~\cite{persephonee2021hotmobile}.
On-device personalisation aims at producing classifiers along the depth of the network that are specialised for the user's data.
At inference time, the model can either exit early if it is confident on its early output, or progressively refine the quality of the result using the deeper exits.
A key advantage of early-exit personalisation is that training can take place even without ground-truth labels in a self-supervised manner, using the output of the DNN's last exit. This implies that a personalised task can become more and more efficient as more personalised inputs are available, without any user supervision. 
Furthermore, personalising only the early exits renders the training process lightweight enough to take place overnight, while the device is plugged in, without the need to access a remote server. This approach was demonstrated by PersEPhonEE~\cite{persephonee2021hotmobile}. By personalising an early-exit \mbox{ResNet-50} using only on-device resources, PersEPhonEE achieved a 2.2$\times$ speedup over the baseline model.

\section{Real-Time AI on Custom Accelerators}
\label{sec:rt-ai-custom-archs}

Towards extracting peak performance and attenuating the sources of inefficiency of standard processors, significant effort has been spent on designing accelerators for DNNs. We define as custom accelerators any architecture that applies domain-specific optimisations~\cite{cnnfpgatoolflows2018csur} and/or approximate computing techniques~\cite{approx_hw2019csur} to trade off lower programmability for higher performance.
Such optimisations can target different components of the underlying hardware. Prominent instances constitute the following.
    
    \textbf{Simplified Control Logic:} The programmable nature of processors requires the use of app-agnostic control logic, which is responsible for tasks such as instruction fetching and accessing the register file.
    Instead, custom accelerators employ a range of techniques to minimise the overhead of this extraneous hardware or replace it with hardwired control. Broadly used techniques include \textit{1)}~domain-specific CISC ISAs and fusion of common operations~\cite{cambricon2016isca,fused2016micro,dnnvm2020tcad} which amortise the overheads of instruction decoding over larger computational work, and \textit{2)}~data-driven streaming execution~\cite{fpgaconvnet2019tnnls,finn2017fpga,latency2017fpl} where processing is triggered whenever data are fed to the accelerator. Such approaches have already been integrated in various accelerators, from Apple's M1 chip~\cite{apple_m1_2020} and Nvidia's Tensor Cores~\cite{tensorcores} to mobile NPUs by Samsung~\cite{samsung_npu2021isca}, Qualcomm~\cite{snpe} and Huawei~\cite{huawei_npu2019hotchips}.

    \textbf{Specialized PE Design:} Representative designs include, but are not limited to, PEs tailored for \textit{i)}~sparse DNNs employing zero-skipping units~\cite{pragmatic2017micro} (\textit{e.g.}~Samsung NPU~\cite{samsung_npu2021isca}), \textit{ii)}~quantised DNNs through custom fixed-~\cite{muppet2020icml} (Qualcomm~\cite{snpe} and Samsung NPUs~\cite{samsung_npu2021isca}) or floating-point representations (\textit{e.g.}~FP16 in Huawei Kirin NPUs~\cite{huawei_npu2019hotchips} and \texttt{ms-fp9} in Microsoft's Brainwave NPU~\cite{brainwave2018isca}, two-precision~\cite{cascadecnn2018fpl,kouris2020date}, mixed-precision~\cite{bitfusion2018isca} (\textit{e.g.}~Nvidia Tensor Cores~\cite{tensorcores}, Qualcomm's 16-bit activations, 8-bit weights (A16W8) in Hexagon 698 NPU~\cite{snpe}) or bit-serial~\cite{stripes2016micro} units, and \textit{ii)}~binarised DNNs (BNNs) with dot-product units replaced with {\small\texttt{popcount}} operators~\cite{finn2017fpga}.
    
    \textbf{Tailored Interconnection:} The inter-PE and PEs-to-buffers interconnect is designed based on the workload of the target DNN~\cite{automated2017dac,maeri2018asplos} for maximum performance and minimum external memory transfers. This is typically driven by the computation-to-communication ratio and the dimensions of the various layers of the target DNN. 
    
    \textbf{Pipeline Organisation:} This comprises accelerators~\cite{fpgaconvnet2019tnnls,streaming2016fpl,finn2017fpga} whose pipelines follow the topology either of the full DNN or of its main building block (\textit{e.g.}~residual block, Inception module, dense block, \textit{etc}). This approach allows the fine-grained allocation of resources among the stages of the pipeline in order to match the processing rate of each stage and reach peak throughput. Similar designs can be found in various commodity devices, such as TV sets with custom AI upscaling processors~\cite{samsung_ai_upscaling2020}.
    
    \textbf{Custom Memory Subsystem:} The on-chip memory organisation is optimised to reduce the external memory bandwidth requirements and increase data-reuse. Such solutions typically restructure the on-chip memory and tailor the buffer sizes to match the DNN workload, while often introducing dedicated compression schemes for weights~\cite{eie2016isca, circcnn2017micro,permdnn2018micro,approx_lstms2020cemag,unzipfpga2021fccm} and feature maps~\cite{eyerissv22019jetcas, def2021aspdac}.

\section{Looking Ahead: The Next Mile in AI Hardware}
\label{sec:future}

Custom hardware is in position to continue being a driving force in providing the computational power and energy efficiency needed for emerging AI-powered consumer platforms.
In this section, we discuss two key directions for AI hardware architectures, namely \textit{i)}~multi-tenant AI accelerators for the concurrent execution of multiple DNNs and \textit{ii)}~automated model-hardware co-design methodologies for the joint optimisation of DNNs and hardware. Furthermore, we discuss how the unique properties of FPGAs can be the key in designing the next-generation of AI processors for consumer devices.

\subsection{Multi-Tenant AI Systems}
\label{sec:multi-tenant-sys}
As the use of AI across applications and users increases, so do the computational demands. In this context, emerging systems tend to employ either pipelines of multiple DNNs or are required to serve queries from different users, each having their own dedicated DNN. This is especially important for inherently multi-tasking platforms, such as smartphones and home robots. However, existing platforms are optimised for the execution of single-DNN apps.
Thus, to cope with this increasing workload, new types of systems have to be developed, specifically optimised for multi-DNN settings.

Mapping multiple DNNs on a computing platform poses important challenges. With each DNN targeting a different task, the performance needs, such as throughput and latency, vary accordingly. This is aggravated by the fact that the multiple DNNs compete for the same pool of resources - off-chip bandwidth and on-chip computational and memory resources. As such, there is an emerging need for solutions that consider both the performance needs of each model and the resource constraints of the underlying platform. Recently, a few works have paved the way towards a new class of multi-DNN systems, encompassing both \textit{1)}~hardware and \textit{2)}~software aspects.

\begin{figure}[t]
    \centering
    {
    \includegraphics[trim={0cm 2cm 0cm 4cm},clip,width=0.45\textwidth]{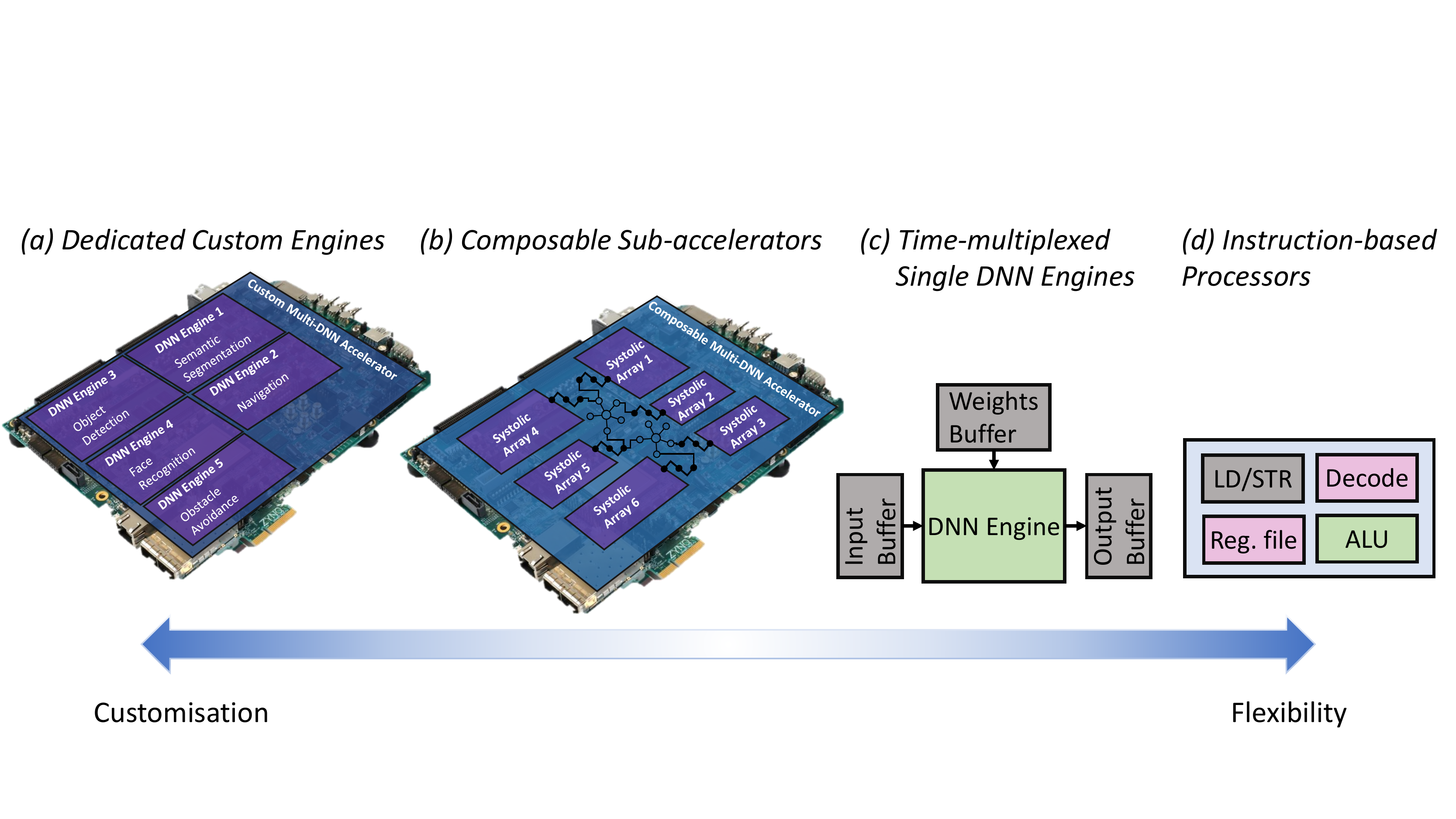}
    \vspace{-0.2cm}
    }
    \caption{Design space of multi-DNN accelerators.}
    \label{fig:multidnn-hw}
     \vspace{-0.65cm}
\end{figure}

\subsubsection{\textbf{Multi-DNN Accelerators}}

Starting from 2018~\cite{fcnnx2018fpl}, a number of accelerators~\cite{fcnnx2018fpl,hda2021hpca,planaria2020micro,aimt2020isca, dataflow_mirroring2021dac,multidnn2020les,infer2020fpt,multilstms2020fpt} have focused on the multi-tenancy scenario. Fig.~\ref{fig:multidnn-hw} shows the spectrum of multi-DNN hardware architectures. Key challenges comprise \textit{i)}~the customisation-programmability trade-off, \textit{i.e.}~how much to customise the hardware for each DNN and how much to reuse across DNNs, and \textit{ii)}~avoiding the resource contention between DNNs, \textit{i.e.}~how to best use the available resources without throttling the performance of the DNNs. The selected strategies for addressing these two issues determine to a great extent the design decisions of the underlying accelerator. 

On the customisation side, f-CNN$^\text{x}$~\cite{fcnnx2018fpl} exploits the static workload of DNN models and derives \textit{dedicated compute engines} for each DNN (Fig.~\ref{fig:multidnn-hw}a), highly tailored to the DNN's workload and application's performance needs. Furthermore, by means of a multi-DNN hardware scheduler, it optimises the external memory bandwidth sharing, in order to minimise the contention between the engines.

Focusing on flexibility, \cite{hda2021hpca} introduces heterogeneous dataflow accelerators (HDAs), which consist of \textit{multiple sub-accelerators} (Fig.~\ref{fig:multidnn-hw}b), each supporting a different dataflow. At run time, each DNN or each DNN layer can be mapped to the most suitable sub-accelerator. 
With the same goal of mapping each DNN layer to the most appropriate engine,
{\small\texttt{Planaria}}~\cite{planaria2020micro} proposes the run-time construction of compute engines by means of multiple composable systolic arrays. Upon execution, the system examines the workloads of the target DNNs, appropriately connects the systolic arrays for each DNN layer and, finally, schedules execution.

With a focus on maximising the resource and bandwidth utilisation, AI-MT~\cite{aimt2020isca} co-locates multiple DNNs on \textit{a single DNN engine} (Fig.~\ref{fig:multidnn-hw}c) and
schedules simultaneously compute- and memory-bound sub-layers of the different DNNs. In this manner, the different sub-layers complementarily utilise the available computational and bandwidth resources, leading to high performance and efficient sharing of the proposed accelerator. 
Similarly, \cite{dataflow_mirroring2021dac} and \cite{prema2020hpca} also target single DNN engines and present dataflow mirroring and a preemption module, respectively, two hardware-level enhancements that aim to optimise the concurrent execution of multiple co-located DNNs on the underlying engine.

Another stream of work investigated the optimal derivation of multi-DNN architectures and the scheduling of DNNs on them through design space exploration~\cite{fcnnx2018fpl,multidnn2020les} and contention-aware performance estimation techniques~\cite{infer2020fpt}.

\begin{figure}[t]
    \centering
    {
    \includegraphics[trim={5cm 10cm 5cm 0cm},clip,width=0.45\textwidth]{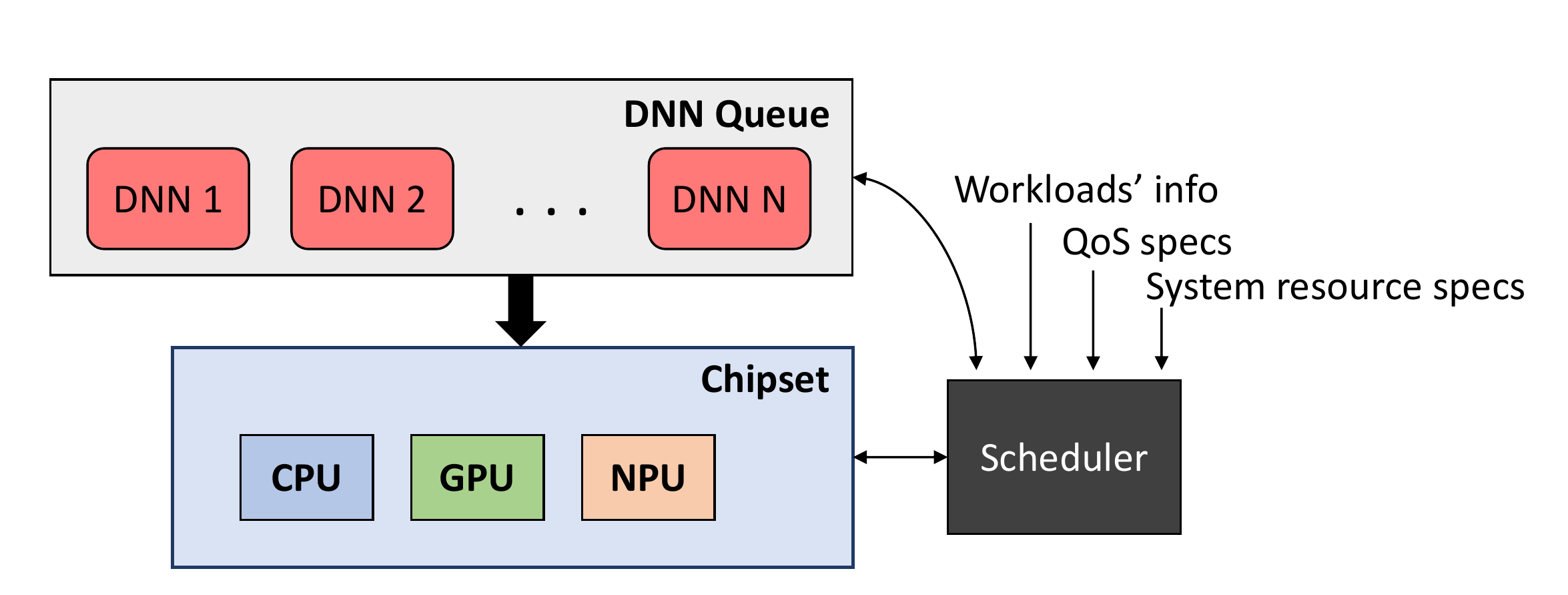}
    \vspace{-0.2cm}
    }
    \caption{System software for multi-DNN systems.}
    \label{fig:multidnn-sw}
     \vspace{-0.65cm}
\end{figure}

\subsubsection{\textbf{Multi-DNN System Software}}
To support multi-DNN application on existing and emerging hardware platforms, a number of software runtimes have been proposed. So far, research effort has been invested on optimising multi-DNN applications for programmable processors (Fig.~\ref{fig:multidnn-hw}d). 
In these works, the \textit{scheduler} (Fig.~\ref{fig:multidnn-sw}) constitutes the most prominent component that effectively determines the attainable performance of the system.
DART~\cite{dart2019rtss} is a scheduler that employs pipelining and priority-based scheduling across heterogeneous processors in order to execute multiple DNN tasks with deterministic response times. PREMA~\cite{prema2020hpca} proposes a DNN-specific priority-based preemptive scheduling algorithm to optimise the execution of multiple models on a single NPU. Similar to AI-MT~\cite{aimt2020isca}, but from a scheduling perspective, \textit{Layerweaver}~\cite{layerweaver2021hpca} introduces a scheme for scheduling together a memory-bound and a compute-bound DNN, improving the utilisation of both the external memory bandwidth and the computational resources. Adopting a different viewpoint, \textsc{Masa}~\cite{masa2021percom} comprises a memory-aware scheduler for minimising the memory swapping between DNNs. With a more model-software co-design approach, NestDNN~\cite{nestdnn2018mobicom} employs multi-capacity models that can dynamically adapt their computational needs. As such, a scheduler can adapt at run time the complexity of each DNN to optimise the overall multi-DNN execution. Finally, targeting mobile robot and IoT platforms, Lee \textit{et al.}~\cite{virt_weights2020mobisys} proposed a weights virtualisation scheme that enables the sharing of weights among DNNs and their efficient in-memory execution.

\subsubsection{\textbf{Open Challenges}}
Here, we discuss open issues and future directions that have only lightly been explored by the initial efforts.

\textbf{Performance vs. Flexibility:}
The early work on multi-DNN systems has currently produced diverse designs with a mostly decoupled consideration of the hardware and software aspects. 
Currently, peak performance is reached through fine-grained customisability~\cite{fcnnx2018fpl} at the expense of a new hardware design cycle whenever a different set of DNNs is targeted. Although this approach may be viable for reconfigurable FPGA-based platforms, where the fabric can be reprogrammed with a different design in the occurrence of a new set of DNNs, ASIC designs require future-proof solutions that can amortise the fabrication cost through broad and efficient re-use across DNN workloads. This performance-flexibility gap is yet to be bridged in the multi-DNN context and remains a promising research avenue.

\textbf{Approximate Computing for Multiple DNNs:}
Another promising approach for exposing more optimisation opportunities for multi-DNN accelerators is approximate computing. Under such schemes, the system would exploit performance-resource usage trade-offs with a controlled drop in accuracy~\cite{approx_hw2019csur}. Examples of such techniques include using different arithmetic precision for each DNN~\cite{kouris2020date} or compressing their weights to a nonuniform degree~\cite{unzipfpga2021fccm}.
For multiple DNNs, this encompasses the development of methods that exploit the cross-DNN redundancy, identify workload commonalities or differences in resilience to quantisation across the DNN models in order to reduce the external memory bandwidth requirements, better coordinate execution and allocate resources among the DNNs. An early approach was presented in \cite{multilstms2020fpt} targeting multi-LSTM applications. In this case, the approximate computing method consists of a parametrised scheme for jointly decomposing the weight matrices of all the target LSTM models, followed by structured pruning and quantisation steps. The design of the associated accelerator is co-optimised together with approximation parameters in order to yield a tailored hardware design that satisfies a user-defined accuracy constraint, leading to 3$\times$-5$\times$ speedup.

\textbf{Multi-DNN Model-Hardware Co-Design:}
Finally, towards extracting both maximum performance and accuracy, model-hardware co-design approaches can be developed that would provide maximal degrees of freedom in the design space. Such methodologies can consider the multiple AI tasks and design from scratch both the DNN architectures and the underlying hardware. 
An early work towards this direction is ASICNAS~\cite{multidnn_codesign2020dac}. To tackle the exponential design space of multi-DNN and accelerator co-optimisation, ASICNAS considers a limited number of pre-defined hardware architectures in its search space. With more than 2$\times$ energy savings and less than 1.6\% accuracy drop, this work showcases the potential of co-design schemes in pushing further the performance of multi-DNN systems. 
Nevertheless, the primary challenge that obstructs multi-DNN model-hardware co-design is still present: the excessively high-dimensional design space that includes model-, scheduling- and hardware-level parameters. As such, research effort needs to be invested in overcoming this complexity through efficient methodologies in order to lead to the next-generation of multi-DNN platforms.

\subsection{Automated Model-Hardware Co-Design}
\label{sec:model-hw-codesign}

Traditional flows in the development of AI products consist of two steps: \textit{1)}~designing and training a DNN model that achieves the required accuracy for the target task under a FLOPs or memory budget and \textit{2)}~optimising the resulting model for execution on a target platforms, \textit{e.g.}~particular mobile phones and IoT devices. In spite of each successes, this approach can lead to suboptimal performance.

An alternative single-stage paradigm that is gaining traction is to \textit{jointly} search for the DNN architecture and the hardware design~\cite{iot_codesgin2019dac,nais2019iccad,best2020dac,hotstart_codesign2020tcad,noc_codesign2020aspdac,hao2021fccm,dance2021dac}. Such a co-design approach can lead to closer-to-optimal configurations and aims to deliver peak performance in terms of both accuracy and processing speed. Nevertheless, main barrier constitutes the excessively large model-hardware design space.  

To counteract the complex design space and explore a sufficiently large number of candidate designs, one line of work~\cite{iot_codesgin2019dac,noc_codesign2020aspdac,best2020dac,codesign2020tcad,hotstart_codesign2020tcad} has adopted pre-defined hardware templates and expose only high-level design parameters in the search space. Others works have incorporated streaming architectures with finer-grained customisability in their hardware design space~\cite{hao2021fccm} or have integrated quantisation into the search space~\cite{hotstart_codesign2020tcad,hao2021fccm}

In an endeavour to push the hardware efficiency to its limits, recent works~\cite{lutnet2020tc,logicnets2020fpl} have designed DNN models that map well to FPGA building blocks. For instance, LUTNet~\cite{lutnet2020tc} and LogicNets~\cite{logicnets2020fpl} incorporate Look-Up Tables (LUTs) as their primitive computational unit, reaching substantial area reduction and throughput gains over both conventional and binarised NNs. The resulting models can be directly mapped to FPGA-based platforms, avoiding the source of inefficiencies of more generic architectures. This is especially important for very resource-constrained platforms in IoT use-cases, where low-cost FPGAs without explicit DSP blocks are often deployed. Nonetheless, with this technology being at its infancy, the high performance currently comes with non-negligible drop in accuracy, which in some applications cannot be tolerated.
As such, to incentivise the wider exploration and potentially real-world adoption of these approaches, the performance and accuracy of such designs has to be scaled up and demonstrated on broader use-cases.

\begin{figure}[t]
    \centering
    {
    \includegraphics[trim={0cm 3cm 4.75cm 5cm},clip,width=0.45\textwidth]{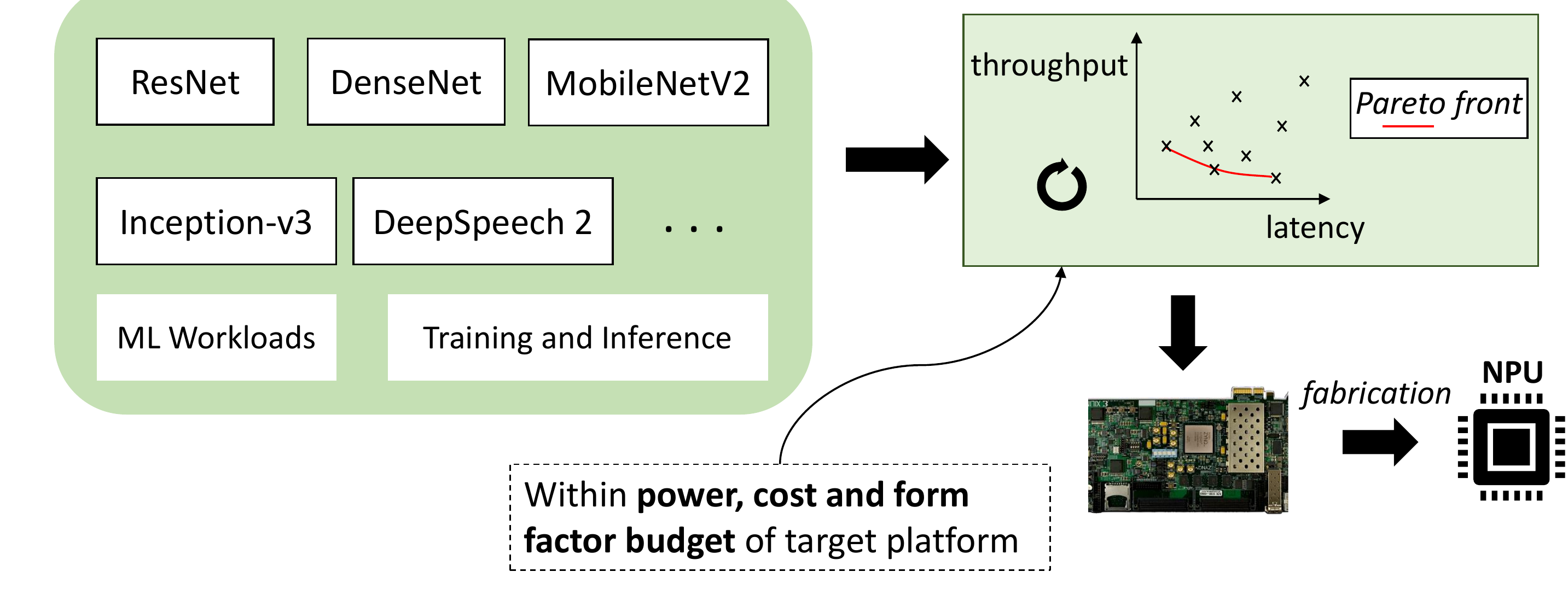}
    \vspace{-0.3cm}
    }
    \caption{FPGA-enabled exploration of next-generation AI processor architectures.}
    \label{fig:npu-search}
     \vspace{-0.65cm}
\end{figure}

\subsection{FPGAs for Deriving Next-Generation AI Processors}
\label{sec:npu-search}

At the moment, there is a constant trend towards integrating NPUs into both mobile SoCs~\cite{embench2019emdl,ai_benchmark2019iccvw} and servers~\cite{tpu2017isca,brainwave2018isca}.
At the same time, deep learning models are evolving rapidly, with architectural changes affecting also their computational characteristics. Due to this, coming up with energy-efficient and high-performance accelerator designs becomes a challenge. In this context, FPGAs can be a key enabler in discovering future NPU designs (Fig.~\ref{fig:npu-search}). By exploiting the reconfigurability of FPGAs, a large number of candidate hardware designs can be explored and run on the FPGA platform to measure critical metrics, including processing speed, power consumption and area.
Given the constraints of the target platform across these dimensions, the objective of this process is to find the Pareto-optimal accelerator design for a number of representative DNN models. 
After the highest performing design has been identified, it can be converted to an ASIC and integrated as an NPU into future consumer devices.

\section{Conclusion}
\label{sec:conclusion}

As real-time AI applications are becoming more and more popular, their use-cases are also becoming more demanding. Supporting such applications on mobile and embedded hardware that is ubiquitous across consumer devices poses important challenges. In this paper, we looked into the current roadblocks that need to be addressed and identified key themes such as the DNN and hardware heterogeneity as well as the dynamicity of the execution environment. Afterwards, we looked into state-of-the-art practices and research directions for both programmable processors and custom accelerators. We further highlighted important future research avenues, with emphasis on multi-tenant inference systems and model-hardware co-design. Our findings reinforce the need to provide solutions across the whole stack; combined research on model, system, platform and hardware optimisations will be of key importance in order to support the next generation of real-time AI applications on mobile/embedded devices.

\bibliographystyle{IEEEtran}
\vspace{-0.2cm}
\bibliography{references.bib}

\begin{thebibliography}{100}
\providecommand{\url}[1]{#1}
\csname url@samestyle\endcsname
\providecommand{\newblock}{\relax}
\providecommand{\bibinfo}[2]{#2}
\providecommand{\BIBentrySTDinterwordspacing}{\spaceskip=0pt\relax}
\providecommand{\BIBentryALTinterwordstretchfactor}{4}
\providecommand{\BIBentryALTinterwordspacing}{\spaceskip=\fontdimen2\font plus
\BIBentryALTinterwordstretchfactor\fontdimen3\font minus
  \fontdimen4\font\relax}
\providecommand{\BIBforeignlanguage}[2]{{%
\expandafter\ifx\csname l@#1\endcsname\relax
\typeout{** WARNING: IEEEtran.bst: No hyphenation pattern has been}%
\typeout{** loaded for the language `#1'. Using the pattern for}%
\typeout{** the default language instead.}%
\else
\language=\csname l@#1\endcsname
\fi
#2}}
\providecommand{\BIBdecl}{\relax}
\BIBdecl

\bibitem{ondevice_asr2020getmobile}
R.~Vipperla \emph{et~al.}, ``{Learning to Listen... On-Device: Present and
  Future Perspectives of On-Device ASR},'' \emph{GetMobile}, 2020.

\bibitem{nvidia_dynamic_sr}
Nvidia, ``{Dynamic Super-Resolution Improves Your Games with 4K-Quality
  Graphics on HD Monitors},'' 2014, [Online; posted 18-September-2014].

\bibitem{neural_enhancement2021csur}
R.~Lee, S.~I. Venieris, and N.~D. Lane, ``{Deep Neural Network-based
  Enhancement for Image and Video Streaming Systems: A Survey and Future
  Directions},'' \emph{ACM Comput. Surv.}, 2021.

\bibitem{complexity}
X.~Xu, Y.~Ding, S.~Xiaobo~Hu, M.~Niemier, J.~Cong, Y.~Hu, and Y.~Shi, ``Scaling
  for edge inference of deep neural networks,'' \emph{Nature Electronics},
  vol.~1, 04 2018.

\bibitem{embench2019emdl}
M.~Almeida \emph{et~al.}, ``{EmBench}: Quantifying performance variations of
  deep neural networks across modern commodity devices,'' in \emph{EMDL}, 2019.

\bibitem{ai_benchmark2019iccvw}
A.~Ignatov \emph{et~al.}, ``{AI Benchmark: All About Deep Learning on
  Smartphones in 2019},'' in \emph{ICCVW}, 2019.

\bibitem{fb_edge2019hpca}
C.~Wu \emph{et~al.}, ``{Machine Learning at Facebook: Understanding Inference
  at the Edge},'' in \emph{HPCA}, 2019.

\bibitem{pruning2020mlsys}
D.~Blalock, J.~J. Gonzalez~Ortiz, J.~Frankle, and J.~Guttag, ``{What is the
  State of Neural Network Pruning?}'' in \emph{MLSys}, 2020.

\bibitem{neurosurgeon2017asplos}
Y.~Kang \emph{et~al.}, ``{Neurosurgeon}: Collaborative intelligence between the
  cloud and mobile edge,'' in \emph{ASPLOS}.

\bibitem{spinn2020mobicom}
S.~Laskaridis \emph{et~al.}, ``{SPINN: Synergistic Progressive Inference of
  Neural Networks over Device and Cloud},'' in \emph{MobiCom}, 2020.

\bibitem{darknetz2020mobisys}
F.~Mo \emph{et~al.}, ``{DarkneTZ: Towards Model Privacy at the Edge Using
  Trusted Execution Environments},'' in \emph{MobiSys}, 2020.

\bibitem{adaptive_mco_2012cloudnet}
{Yuan Zhang}, {Hao Liu}, {Lei Jiao}, and {Xiaoming Fu}, ``{To offload or not to
  offload: An efficient code partition algorithm for mobile cloud computing},''
  in \emph{CLOUDNET}, 2012.

\bibitem{self_driving_cars2018asplos}
S.-C. Lin \emph{et~al.}, ``{The Architectural Implications of Autonomous
  Driving: Constraints and Acceleration},'' in \emph{ASPLOS}, 2018.

\bibitem{vr_latency2018mobisys}
L.~Liu \emph{et~al.}, ``{Cutting the Cord: Designing a High-Quality Untethered
  VR System with Low Latency Remote Rendering},'' in \emph{MobiSys}, 2018.

\bibitem{approx_hw2019csur}
E.~Wang, J.~J. Davis, R.~Zhao, H.-C. Ng, X.~Niu, W.~Luk, P.~Y.~K. Cheung, and
  G.~A. Constantinides, ``{Deep Neural Network Approximation for Custom
  Hardware: Where We've Been, Where We're Going},'' \emph{ACM Comput. Surv.},
  2019.

\bibitem{mobisr2019mobicom}
R.~Lee, S.~I. Venieris, L.~Dudziak, S.~Bhattacharya, and N.~D. Lane, ``{MobiSR:
  Efficient On-Device Super-Resolution through Heterogeneous Mobile
  Processors},'' in \emph{MobiCom}, 2019.

\bibitem{mess2021arxiv}
A.~Kouris, S.~I. Venieris, S.~Laskaridis, and N.~D. Lane, ``{Multi-Exit
  Semantic Segmentation Networks},'' in \emph{arXiv}, 2021.

\bibitem{tpu2017isca}
N.~P. Jouppi \emph{et~al.}, ``{In-Datacenter Performance Analysis of a Tensor
  Processing Unit},'' in \emph{ISCA}, 2017.

\bibitem{brainwave2018isca}
J.~Fowers \emph{et~al.}, ``{A Configurable Cloud-Scale DNN Processor for
  Real-Time AI},'' in \emph{ISCA}, 2018.

\bibitem{nas2017iclr}
B.~Zoph and Q.~V. Le, ``{Neural Architecture Search with Reinforcement
  Learning},'' in \emph{ICLR}, 2017.

\bibitem{rwn2019iccv}
S.~Xie, A.~Kirillov, R.~Girshick, and K.~He, ``{Exploring Randomly Wired Neural
  Networks for Image Recognition},'' in \emph{ICCV}, 2019.

\bibitem{rwn2019neurips}
M.~Wortsman, A.~Farhadi, and M.~Rastegari, ``{Discovering Neural Wirings},'' in
  \emph{NeurIPS}, 2019.

\bibitem{rwn2020mlsys}
B.~H. Ahn \emph{et~al.}, ``{Ordering Chaos: Memory-Aware Scheduling of
  Irregularly Wired Neural Networks for Edge Devices},'' in \emph{MLSys}, 2020.

\bibitem{rwn2020fpl}
R.~Kuramochi and H.~Nakahara, ``{An FPGA-Based Low-Latency Accelerator for
  Randomly Wired Neural Networks},'' in \emph{FPL}, 2020.

\bibitem{oodin2021smartcomp}
S.~I. Venieris, I.~Panopoulos, and I.~S. Venieris, ``{OODIn: An Optimised
  On-Device Inference Framework for Heterogeneous Mobile Devices},'' in
  \emph{IEEE SMARTCOMP}, 2021.

\bibitem{nestdnn2018mobicom}
B.~Fang \emph{et~al.}, ``{NestDNN: Resource-Aware Multi-Tenant On-Device Deep
  Learning for Continuous Mobile Vision},'' in \emph{MobiCom}, 2018.

\bibitem{dvfs2020mdat}
A.~K. Singh \emph{et~al.}, ``{Dynamic Energy and Thermal Management of
  Multi-core Mobile Platforms: A Survey},'' \emph{IEEE Design Test}, 2020.

\bibitem{realitycheck2019edgesys}
A.~Cartas \emph{et~al.}, ``{A Reality Check on Inference at Mobile Networks
  Edge},'' in \emph{EdgeSys}, 2019.

\bibitem{exynos_cpu2020isca}
B.~Grayson \emph{et~al.}, ``{Evolution of the Samsung Exynos CPU
  Microarchitecture},'' in \emph{ISCA}, 2020.

\bibitem{proc_inefficiencies2010isca}
R.~Hameed, W.~Qadeer, M.~Wachs, O.~Azizi, A.~Solomatnikov, B.~C. Lee,
  S.~Richardson, C.~Kozyrakis, and M.~Horowitz, ``{Understanding Sources of
  Inefficiency in General-Purpose Chips},'' in \emph{ISCA}, 2010.

\bibitem{nimble2021mlsys}
H.~Shen \emph{et~al.}, ``{Nimble: Efficiently Compiling Dynamic Neural Networks
  for Model Inference},'' in \emph{MLSys}, 2021.

\bibitem{adaptive_dnns2021emdl}
S.~Laskaridis, A.~Kouris, and N.~D. Lane, ``{Adaptive Inference through
  Early-Exit Networks: Design, Challenges and Directions},'' in \emph{EMDL},
  2021.

\bibitem{dyno2021arxiv}
M.~Almeida, S.~Laskaridis, S.~I. Venieris, I.~Leontiadis, and N.~D. Lane,
  ``{DynO: Dynamic Onloading of Deep Neural Networks from Cloud to Device},''
  in \emph{arXiv}, 2021.

\bibitem{integer_dnn2018cvpr}
B.~Jacob \emph{et~al.}, ``{Quantization and Training of Neural Networks for
  Efficient Integer-Arithmetic-Only Inference},'' in \emph{CVPR}, 2018.

\bibitem{netadapt2018eccv}
T.-J. Yang \emph{et~al.}, ``{NetAdapt: Platform-Aware Neural Network Adaptation
  for Mobile Applications},'' in \emph{ECCV}, 2018.

\bibitem{approx_lstms2020cemag}
A.~Kouris, S.~I. Venieris, M.~Rizakis, and C.-S. Bouganis, ``{Approximate LSTMs
  for Time-Constrained Inference: Enabling Fast Reaction in Self-Driving
  Cars},'' \emph{IEEE Consumer Electronics Magazine}, 2020.

\bibitem{dyn_channel_pruning2018iclr}
X.~Gao, Y.~Zhao, {\L}.~Dudziak, R.~Mullins, and C.-z. Xu, ``{Dynamic Channel
  Pruning: Feature Boosting and Suppression},'' in \emph{ICLR}, 2018.

\bibitem{hapi2020iccad}
S.~Laskaridis, S.~I. Venieris, H.~Kim, and N.~D. Lane, ``{HAPI: Hardware-Aware
  Progressive Inference},'' in \emph{ICCAD}, 2020.

\bibitem{brpnas2020neurips}
L.~Dudziak, T.~Chau, M.~Abdelfattah, R.~Lee, H.~Kim, and N.~Lane, ``{BRP-NAS:
  Prediction-based NAS using GCNs},'' in \emph{NeurIPS}, 2020.

\bibitem{mobilenetv2_2018cvpr}
M.~Sandler \emph{et~al.}, ``{MobileNetV2: Inverted Residuals and Linear
  Bottlenecks},'' in \emph{CVPR}, 2018.

\bibitem{squeezenet2016arxiv}
F.~N. Iandola, S.~Han, M.~W. Moskewicz, K.~Ashraf, W.~J. Dally, and K.~Keutzer,
  ``{SqueezeNet: AlexNet-level accuracy with 50x fewer parameters and $<$ 0.5
  MB model size},'' \emph{arXiv}, 2016.

\bibitem{efficientnet2019icml}
M.~Tan and Q.~Le, ``{EfficientNet: Rethinking Model Scaling for Convolutional
  Neural Networks},'' in \emph{ICML}, 2019.

\bibitem{morphnet2018cvpr}
A.~Gordon \emph{et~al.}, ``{MorphNet: Fast Simple Resource-Constrained
  Structure Learning of Deep Networks},'' in \emph{CVPR}, 2018.

\bibitem{tpsr2020eccv}
R.~Lee, {\L}.~Dudziak, M.~Abdelfattah, S.~I. Venieris, H.~Kim, H.~Wen, and
  N.~D. Lane, ``{Journey Towards Tiny Perceptual Super-Resolution},'' in
  \emph{ECCV}, 2020.

\bibitem{shrinkml2019interspeech}
Łukasz Dudziak, M.~S. Abdelfattah, R.~Vipperla, S.~Laskaridis, and N.~D. Lane,
  ``{ShrinkML: End-to-End ASR Model Compression Using Reinforcement
  Learning},'' in \emph{Interspeech}, 2019.

\bibitem{compressed_asr2020interspeech}
A.~Mehrotra \emph{et~al.}, ``{Iterative Compression of End-to-End ASR Model
  Using AutoML},'' in \emph{Interspeech}, 2020.

\bibitem{bunched_tts2020intespeech}
R.~Vipperla \emph{et~al.}, ``{Bunched LPCNet: Vocoder for Low-Cost Neural
  Text-To-Speech Systems},'' in \emph{Interspeech}, 2020.

\bibitem{persephonee2021hotmobile}
I.~Leontiadis, S.~Laskaridis, S.~I. Venieris, and N.~D. Lane, ``{It's Always
  Personal: Using Early Exits for Efficient On-Device CNN Personalisation},''
  in \emph{HotMobile}, 2021.

\bibitem{cnnfpgatoolflows2018csur}
S.~I. Venieris, A.~Kouris, and C.-S. Bouganis, ``{Toolflows for Mapping
  Convolutional Neural Networks on FPGAs: A Survey and Future Directions},''
  \emph{ACM Comput. Surv.}, 2018.

\bibitem{cambricon2016isca}
S.~Liu \emph{et~al.}, ``{Cambricon: An Instruction Set Architecture for Neural
  Networks},'' in \emph{ISCA}, 2016.

\bibitem{fused2016micro}
M.~Alwani, H.~Chen, M.~Ferdman, and P.~Milder, ``{Fused-Layer CNN
  Accelerators},'' in \emph{MICRO}, 2016.

\bibitem{dnnvm2020tcad}
Y.~Xing \emph{et~al.}, ``{DNNVM}: End-to-end compiler leveraging heterogeneous
  optimizations on {FPGA}-based {CNN} accelerators,'' \emph{TCAD}, 2020.

\bibitem{fpgaconvnet2019tnnls}
S.~I. Venieris and C.-S. Bouganis, ``{fpgaConvNet: Mapping Regular and
  Irregular Convolutional Neural Networks on FPGAs},'' \emph{TNNLS}, 2019.

\bibitem{finn2017fpga}
Y.~Umuroglu \emph{et~al.}, ``{FINN: A Framework for Fast, Scalable Binarized
  Neural Network Inference},'' in \emph{FPGA}, 2017.

\bibitem{latency2017fpl}
S.~I. Venieris and C.-S. Bouganis, ``{Latency-Driven Design for FPGA-based
  Convolutional Neural Networks},'' in \emph{FPL}, 2017.

\bibitem{apple_m1_2020}
Apple, ``{Apple M1},''
  \url{https://www.apple.com/newsroom/2020/11/apple-unleashes-m1/}, 2020,
  accessed: \today.

\bibitem{tensorcores}
\BIBentryALTinterwordspacing
J.~Appleyard and S.~Yokim, ``{Programming Tensor Cores in CUDA 9},'' October
  2017, [Online; posted 17-October-2017]. [Online]. Available:
  \url{https://devblogs.nvidia.com/programming-tensor-cores-cuda-9/}
\BIBentrySTDinterwordspacing

\bibitem{samsung_npu2021isca}
J.-W. Jang \emph{et~al.}, ``{Sparsity-Aware and Re-configurable NPU
  Architecture for Samsung Flagship Mobile SoC},'' in \emph{ISCA}, 2021.

\bibitem{snpe}
Qualcomm, ``{Snapdragon Neural Processing Engine},''
  \url{https://developer.qualcomm.com/docs/snpe/snapdragon\_npe\_runtime.html},
  2021, accessed: \today.

\bibitem{huawei_npu2019hotchips}
H.~{Liao}, J.~{Tu}, J.~{Xia}, and X.~{Zhou}, ``{DaVinci: A Scalable
  Architecture for Neural Network Computing},'' in \emph{HotChips}, 2019, pp.
  1--44.

\bibitem{pragmatic2017micro}
J.~Albericio \emph{et~al.}, ``{Bit-Pragmatic Deep Neural Network Computing},''
  in \emph{MICRO}, 2017.

\bibitem{muppet2020icml}
A.~Rajagopal, D.~Vink, S.~Venieris, and C.-S. Bouganis, ``{Multi-Precision
  Policy Enforced Training ({M}u{PPET}) : A Precision-Switching Strategy for
  Quantised Fixed-Point Training of {CNN}s},'' in \emph{ICML}, 2020.

\bibitem{cascadecnn2018fpl}
A.~Kouris, S.~I. Venieris, and C.-S. Bouganis, ``{CascadeCNN: Pushing the
  Performance Limits of Quantisation in Convolutional Neural Networks},'' in
  \emph{FPL}, 2018.

\bibitem{kouris2020date}
------, ``{A Throughput-Latency Co-Optimised Cascade of Convolutional Neural
  Network Classifiers},'' in \emph{DATE}, 2020.

\bibitem{bitfusion2018isca}
H.~Sharma \emph{et~al.}, ``{Bit Fusion: Bit-Level Dynamically Composable
  Architecture for Accelerating Deep Neural Network},'' in \emph{ISCA}, 2018.

\bibitem{stripes2016micro}
P.~Judd \emph{et~al.}, ``{Stripes: Bit-Serial Deep Neural Network Computing},''
  in \emph{MICRO}, 2016.

\bibitem{automated2017dac}
X.~Wei, C.~H. Yu, P.~Zhang, Y.~Chen, Y.~Wang, H.~Hu, Y.~Liang, and J.~Cong,
  ``{Automated Systolic Array Architecture Synthesis for High Throughput CNN
  Inference on FPGAs},'' in \emph{DAC}, 2017.

\bibitem{maeri2018asplos}
H.~Kwon, A.~Samajdar, and T.~Krishna, ``{MAERI: Enabling Flexible Dataflow
  Mapping over DNN Accelerators via Reconfigurable Interconnects},'' in
  \emph{ASPLOS}, 2018.

\bibitem{streaming2016fpl}
{Huimin Li} \emph{et~al.}, ``{A High Performance FPGA-based Accelerator for
  Large-Scale Convolutional Neural Networks},'' in \emph{FPL}, 2016.

\bibitem{samsung_ai_upscaling2020}
Samsung, ``{AI Upscaling on Samsung TVs},''
  \url{https://www.samsung.com/au/support/tv-audio-video/ai-upscaling-on-samsung-tvs/},
  2020, accessed: \today.

\bibitem{eie2016isca}
S.~{Han} \emph{et~al.}, ``{EIE: Efficient Inference Engine on Compressed Deep
  Neural Network},'' in \emph{ISCA}, 2016.

\bibitem{circcnn2017micro}
C.~{Ding} \emph{et~al.}, ``{CirCNN: Accelerating and Compressing Deep Neural
  Networks Using Block-Circulant Weight Matrices},'' in \emph{MICRO}, 2017.

\bibitem{permdnn2018micro}
C.~Deng \emph{et~al.}, ``{PermDNN: Efficient Compressed DNN Architecture with
  Permuted Diagonal Matrices},'' in \emph{MICRO}, 2018.

\bibitem{unzipfpga2021fccm}
S.~I. Venieris, J.~Fernandez-Marques, and N.~D. Lane, ``{unzipFPGA: Enhancing
  FPGA-based CNN Engines with On-the-Fly Weights Generation},'' in \emph{FCCM},
  2021.

\bibitem{eyerissv22019jetcas}
Y.~{Chen}, T.~{Yang}, J.~{Emer}, and V.~{Sze}, ``{Eyeriss v2: A Flexible
  Accelerator for Emerging Deep Neural Networks on Mobile Devices},''
  \emph{JETCAS}, 2019.

\bibitem{def2021aspdac}
A.~Montgomerie-Corcoran and C.-S. Bouganis, ``{DEF: Differential Encoding of
  Featuremaps for Low Power Convolutional Neural Network Accelerators},'' in
  \emph{ASP-DAC}, 2021.

\bibitem{fcnnx2018fpl}
S.~I. Venieris and C.-S. Bouganis, ``{f-CNNx: A Toolflow for Mapping Multiple
  Convolutional Neural Networks on FPGAs},'' in \emph{FPL}, 2018.

\bibitem{hda2021hpca}
H.~Kwon \emph{et~al.}, ``{Heterogeneous Dataflow Accelerators for Multi-DNN
  Workloads},'' in \emph{HPCA}, 2021.

\bibitem{planaria2020micro}
S.~Ghodrati \emph{et~al.}, ``{Planaria}: Dynamic architecture fission for
  spatial multi-tenant acceleration of deep neural networks,'' in \emph{MICRO},
  2020.

\bibitem{aimt2020isca}
E.~Baek, D.~Kwon, and J.~Kim, ``{A Multi-Neural Network Acceleration
  Architecture},'' in \emph{ISCA}, 2020.

\bibitem{dataflow_mirroring2021dac}
J.~Lee, J.~Choi, J.~Kim, J.~Lee, and Y.~Kim, ``{Dataflow Mirroring:
  Architectural Support for Highly Efficient Fine-Grained Spatial Multitasking
  on Systolic-Array NPUs},'' in \emph{DAC}, 2021.

\bibitem{multidnn2020les}
R.~Kedia, S.~Goel, M.~Balakrishnan, K.~Paul, and R.~Sen, ``{Design Space
  Exploration of FPGA Based System with Multiple DNN Accelerators},''
  \emph{IEEE Embedded Systems Letters}, 2020.

\bibitem{infer2020fpt}
S.~Goel, R.~Kedia, M.~Balakrishnan, and R.~Sen, ``{INFER: INterFerence-aware
  Estimation of Runtime for Concurrent CNN Execution on DPUs},'' in
  \emph{ICFPT}, 2020.

\bibitem{multilstms2020fpt}
S.~Ribes, P.~Trancoso, I.~Sourdis, and C.-S. Bouganis, ``{Mapping Multiple LSTM
  models on FPGAs},'' in \emph{ICFPT}, 2020.

\bibitem{prema2020hpca}
Y.~Choi and M.~Rhu, ``{PREMA: A Predictive Multi-Task Scheduling Algorithm for
  Preemptible Neural Processing Units},'' in \emph{HPCA}, 2020.

\bibitem{dart2019rtss}
Y.~Xiang and H.~Kim, ``{Pipelined Data-Parallel CPU/GPU Scheduling for
  Multi-DNN Real-Time Inference},'' in \emph{RTSS}, 2019.

\bibitem{layerweaver2021hpca}
Y.~H. Oh \emph{et~al.}, ``{Layerweaver: Maximizing Resource Utilization of
  Neural Processing Units via Layer-Wise Scheduling},'' in \emph{HPCA}, 2021.

\bibitem{masa2021percom}
B.~Cox, J.~Galjaard, A.~Ghiassi, R.~Birke, and L.~Y. Chen, ``{Masa: Responsive
  Multi-DNN Inference on the Edge},'' in \emph{PerCom}, 2021.

\bibitem{virt_weights2020mobisys}
S.~Lee and S.~Nirjon, ``{Fast and Scalable In-Memory Deep Multitask Learning
  via Neural Weight Virtualization},'' in \emph{MobiSys}, 2020.

\bibitem{multidnn_codesign2020dac}
L.~Yang \emph{et~al.}, ``{Co-Exploration of Neural Architectures and
  Heterogeneous ASIC Accelerator Designs Targeting Multiple Tasks},'' in
  \emph{DAC}, 2020.

\bibitem{iot_codesgin2019dac}
C.~Hao \emph{et~al.}, ``{FPGA/DNN Co-Design: An Efficient Design Methodology
  for IoT Intelligence on the Edge},'' in \emph{DAC}, 2019.

\bibitem{nais2019iccad}
C.~Hao, Y.~Chen \emph{et~al.}, ``{NAIS: Neural Architecture and Implementation
  Search and its Applications in Autonomous Driving},'' in \emph{ICCAD}, 2019.

\bibitem{best2020dac}
M.~S. Abdelfattah, {\L}.~Dudziak, T.~Chau, R.~Lee, H.~Kim, and N.~D. Lane,
  ``{Best of Both Worlds: AutoML Codesign of a CNN and its Hardware
  Accelerator},'' in \emph{DAC}, 2020.

\bibitem{hotstart_codesign2020tcad}
W.~Jiang \emph{et~al.}, ``Standing on the shoulders of giants: Hardware and
  neural architecture co-search with hot start,'' \emph{TCAD}.

\bibitem{noc_codesign2020aspdac}
L.~Yang \emph{et~al.}, ``Co-exploring neural architecture and network-on-chip
  design for real-time artificial intelligence,'' in \emph{ASP-DAC}.

\bibitem{hao2021fccm}
Z.~Dong \emph{et~al.}, ``{HAO: Hardware-aware Neural Architecture Optimization
  for Efficient Inference},'' in \emph{FCCM}, 2021.

\bibitem{dance2021dac}
K.~Choi, D.~Hong, H.~Yoon, J.~Yu, Y.~Kim, and J.~Lee, ``{DANCE: Differentiable
  Accelerator/Network Co-Exploration},'' in \emph{DAC}, 2021.

\bibitem{codesign2020tcad}
W.~Jiang \emph{et~al.}, ``{Hardware/Software Co-Exploration of Neural
  Architectures},'' \emph{TCAD}, 2020.

\bibitem{lutnet2020tc}
E.~Wang, J.~J. Davis, P.~Y.~K. Cheung, and G.~A. Constantinides, ``{LUTNet:
  Learning FPGA Configurations for Highly Efficient Neural Network
  Inference},'' \emph{TC}, 2020.

\bibitem{logicnets2020fpl}
Y.~{Umuroglu} \emph{et~al.}, ``{LogicNets: Co-Designed Neural Networks and
  Circuits for Extreme-Throughput Applications},'' in \emph{FPL}, 2020.

\end{thebibliography}

\end{document}